%% file: main.tex
\documentclass[letterpaper, 10pt, twocolumn]{IEEEtran}

\usepackage{amsmath,amssymb,amsthm}
\usepackage{bm}
\usepackage{booktabs}
\usepackage{tabularx}
\usepackage{algorithm}
\usepackage{algpseudocode}
\usepackage{tikz}
\usetikzlibrary{arrows.meta, positioning, shapes, calc, fit}
\usepackage{pgfplots}
\pgfplotsset{compat=1.17}
\usepackage{graphicx}
\usepackage[pagebackref=false, colorlinks=true, citecolor=blue, linkcolor=blue, urlcolor=blue]{hyperref}
\usepackage{acronym}

\usepackage[none]{comment} 

\usepackage[noadjust]{cite}

\graphicspath{{figures/}}

\theoremstyle{definition}

\algnewcommand{\Continue}{\textbf{continue}}
\algnewcommand{\Break}{\textbf{break}}

\begin{document}

\title{LiDAR-based Humanoid Navigation in Dynamic, Non-Convex Environments via Spatio-Temporal Safe Corridors}

\author{Shubham~S.~Kumbhar
       and~Panagiotis~Artemiadis$^*$, \textit{IEEE Senior Member}


\thanks{Shubham S. Kumbhar, and Panagiotis Artemiadis are with the Mechanical Engineering Department, at the University of Delaware, Newark, DE 19716, USA. {\tt\small shubhamk@udel.edu, partem@udel.edu}}
\thanks{$^*$Corrresponding author}}%


\maketitle

\begin{abstract}

Navigating humanoid robots through dynamic, non-convex environments requires tightly coupling onboard perception with footstep planning and dynamic locomotion control. Existing methods frequently discard complex free space through simplified convex obstacle bounds and fail to account for real-time dynamic occupancy predictions. 
This paper presents a real-time framework for humanoid navigation that directly constructs spatio-temporal safe corridors from LiDAR-based particle representations and couples them with Linear Inverted Pendulum (LIP)-based planning and Whole Body Control (WBC) to enable navigation around and within moving non-convex structures without restrictive geometric assumptions.
Simulations on the Digit humanoid yield a 96\% success rate in navigating around and inside moving non-convex obstacles in a $\mathbf{24\times24~m^2}$ arena. Ablation studies confirm that the proposed iterative polytope generation significantly outperforms single-iteration baselines in narrow corridors, while evaluations demonstrate robustness to clutter, sensor noise, control delays, and force perturbations.
\end{abstract}


\input{sections/01_introduction}

\input{sections/02_preliminaries}

\input{sections/03_methodology}
\input{sections/04_experiments}

\input{sections/05_conclusion}

\bibliographystyle{IEEEtran}
\bibliography{refs}

\end{document}

%% file: sections/01_introduction.tex
\section{Introduction}
Humanoid robots have the potential to operate alongside humans in cluttered, dynamic environments, yet autonomous navigation remains challenging for bipedal systems. Unlike wheeled platforms, bipedal robots regulate their center-of-mass (CoM) dynamics through discrete foot contacts, coupling collision avoidance with the selection and timing of feasible footsteps. Model-based approaches, particularly reduced-order model-based Model Predictive Control (MPC) and Quadratic Programming (QP), have enabled dynamically stable bipedal locomotion \cite{Poulakakis2022Sequential, Kumbhar2025MPC, Ding2022Orientation, Nguyen2015Safety}, motivating their use in obstacle-aware navigation.


Static-environment navigation was explored in \cite{simone2017mpc} using half-plane constraints in MPC, assuming known obstacle geometry and position. An end-to-end quadruped navigation framework in \cite{gilroy2021autonomous} combined vision-based mapping and global $A^*$ planning with walking and jumping over static obstacles, but did not address dynamic obstacle avoidance. Safe-corridor-based navigation was explored in \cite{Poulakakis2022Sequential}, which incorporated dynamic obstacle avoidance constraints into MPC but assumed simplified convex obstacle geometries and direct availability of obstacle position and velocity information.

The advancements associated with static obstacle navigation have naturally motivated extensions to dynamic environments, where obstacle motion must also be considered. A perception-to-walking pipeline for the humanoid Lola using Swept-Sphere-Volume (SSV) primitives \cite{wahrmann2016fast} was developed in \cite{hildebrandt2015real} and later extended with hierarchical footstep replanning \cite{hildebrandt2017real}. While this representation can capture non-convex obstacles, the framework operates reactively without explicitly predicting obstacle motion. Similarly, a reactive avoidance scheme using spherical obstacle representations was employed in \cite{rossini2023real}. MPC-QP-based avoidance was extended to fast-moving objects in \cite{wang2025dynamic} using half-plane constraints and velocity-stretched ellipsoids for obstacles. However, that approach treats these ellipsoids as static over the MPC horizon rather than incorporating time-varying obstacle predictions, and assumes a priori availability of obstacle states.

Predictive MPC-QP obstacle avoidance using half-space relaxations of predicted obstacle motion was introduced in \cite{white2022avoiding}. This approach was extended to bipedal standing and stepping avoidance in \cite{wang2023real}. However, both approaches rely on simplified obstacle geometries and externally provided obstacle states. Additionally, these methods are not designed for global navigation and may get trapped in local minima.

Social Zonotope Network (SZN)-MPC was proposed in \cite{shamsah2024socially} for avoiding dynamic obstacles in social navigation scenarios. Specifically, it predicts pedestrian motion using convex zonotopes for MPC-based collision avoidance. However, this representation primarily captures motion uncertainty rather than the physical geometry of obstacles. A learning-based approach, the reactive Agile But Safe (ABS) framework, was proposed in \cite{he2024agile} for high-speed quadrupedal locomotion, while the Reflexive Evasion Robot (REBot) framework was proposed in \cite{xu2025rebot} for high-speed evasion. Both approaches remain primarily local and rely on simplified sensing or obstacle representations, limiting their applicability to long-horizon navigation and explicit geometry-aware safety.

Together, the existing approaches reveal two key limitations. First, obstacle geometry is often simplified using spheres or ellipsoids, which can conservatively eliminate free space around non-convex obstacles. This can fundamentally restrict achievable navigation tasks. For example, a convex approximation of a C- or U-shaped obstacle may classify its concave interior as occupied, preventing a humanoid from navigating within the enclosed region. Second, most model-based approaches assume that obstacle geometry, position, and/or velocity are already available. Consequently, only a few approaches connect onboard spatial sensing and dynamic occupancy prediction with geometry-aware free-space construction, footstep selection, and dynamically stable locomotion.

Dynamic obstacle avoidance has been explored more extensively in aerial robotics, where full-stack perception, prediction, planning, and control systems have been developed. 
Particle-based dynamic occupancy representations such as the Dual-Structure Particle (DSP) map enable complex obstacle geometries to be represented while maintaining a lightweight structure suitable for real-time prediction. It was utilized in \cite{chen2022rast} to construct risk-aware spatio-temporal (RAST) safety corridors for drone navigation.
However, its rectangular corridor representation can remain conservative around complex obstacle boundaries and does not account for the discrete footstep selection and dynamic stability constraints inherent to bipedal locomotion.

In this work, we present an algorithmic systems framework that brings the advantages of particle-based dynamic occupancy representations to bipedal navigation and tightly integrates them with footstep planning and locomotion control. The primary contributions are: (i) a \textbf{full-stack perception-to-control framework for humanoid navigation} that bridges 3D Light Detection and Ranging (LiDAR) perception and dynamic locomotion control through a modified DSP-map representation, A* global guidance, capture-region-constrained Linear Inverted Pendulum (LIP) sampling, LIP-MPC, and WBC, enabling collision-free navigation in dynamic, non-convex environments; (ii) a \textbf{spatio-temporal safe corridor generation method} that uses capture-region-constrained LIP trajectory sampling to extract kinematically and dynamically feasible seed points and a risk-aware iterative polytope generation method with per-cluster risk thresholds to construct large, temporally consistent convex free regions from predicted particle-based obstacle occupancy; and (iii) \textbf{comprehensive simulation-based validation} of the complete framework on the Digit humanoid in dynamic, non-convex environments, including navigation within moving non-convex obstacles and robustness to sim-to-real imperfections such as sensor noise, control delays, and external perturbations.

The remainder of this paper is organized as follows: Section~\ref{sec:perception} details the $360^\circ$ LiDAR perception and the modified DSP-map. Section~\ref{sec:planning&control} presents the planning and control pipeline, covering A*-based global guidance, LIP trajectory sampling, safe-polytope generation, and MPC-WBC execution. Section~\ref{sec:results} provides simulation results and Section~\ref{sec:conclusion} concludes the paper.

\begin{figure}[t]
    \centering
    \includegraphics[width=0.48\textwidth]{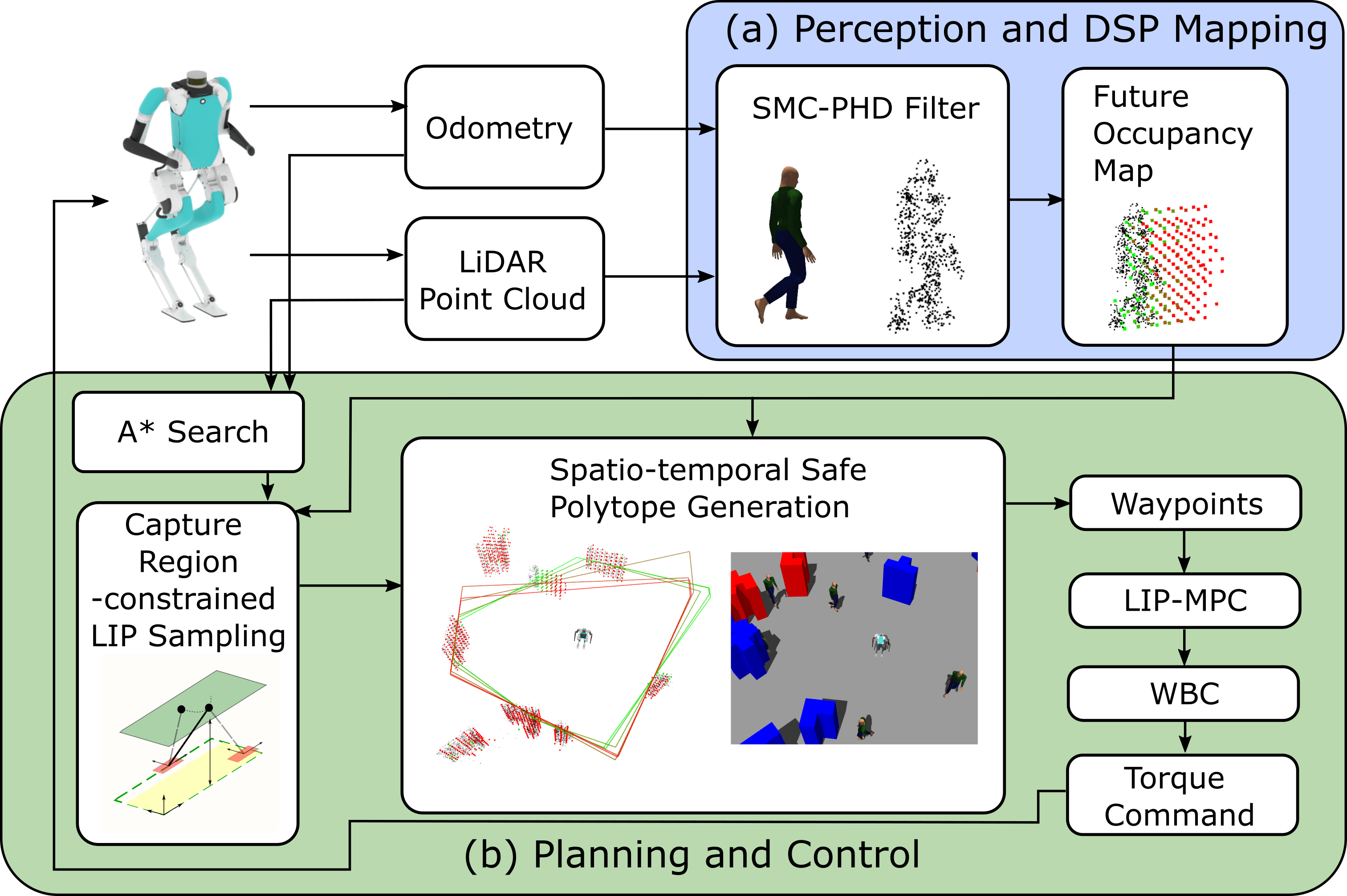}
    \vspace{-0.3cm}
    \caption{Proposed Perception-to-Control Framework. (a) Perception module and DSP Mapping for particle-based obstacle representation and motion prediction. (b) Planning and control module for global navigation.}
    \vspace{-0.6cm}
    \label{fig:pipeline}
\end{figure}



%% file: sections/02_preliminaries.tex
\section{Perception Module}
\label{sec:perception}

The DSP map~\cite{chen2023continuous} is an egocentric local mapping framework that represents arbitrarily shaped static and dynamic obstacles as a set of weighted particles using Random Finite Set (RFS) theory. Each particle $(i)$ represents an obstacle state ($\tilde{\mathbf{x}}_{t_0}^{(i)}$) comprising its 3D position ($\tilde{\mathbf{p}}_{t_0}^{(i)}$) and velocity ($\tilde{\mathbf{v}}_{t_0}^{(i)}$),
with an associated weight $w_{t_0}^{(i)}$ at time $t_0$. The particle weights represent the local intensity of obstacle point objects, with higher weights indicating a greater contribution to the estimated spatial occupancy. The environment is represented by the Probability Hypothesis Density (PHD) of the random finite set of the particles. We denote this PHD by $D(\mathbf{x}_{t_0})$, which is approximated using a Sequential Monte Carlo PHD (SMC-PHD) filter as
\begin{equation}
D(\mathbf{x}_{t_0}) =
\sum_{i=1}^{n_{t_0}} w_{t_0}^{(i)}
\delta\left(\mathbf{x}_{t_0}-\tilde{\mathbf{x}}_{t_0}^{(i)}\right).
\end{equation}
where $\delta(.)$ is the Dirac function, $n_{t_0}$ is the cardinality of the RFS of the particles and $\mathbf{x}_{t_0}\in\mathbb{R}^3$ is a point in 3D space.
The filter updates particle states and weights from depth-sensor point clouds and robot odometry, thereby providing a local representation that captures both obstacle geometry and motion. In this work, we extend the DSP map with the modifications described below to support LiDAR-based perception and improve prediction quality for humanoid navigation.
\vspace{-0.3cm}
\subsection{Modifications to the DSP Map}
\paragraph{LiDAR Extension}
 
The original DSP map \cite{chen2023continuous} was developed for depth sensors with a limited field of view and uses pyramid-like subspace division for efficient particle propagation and occlusion handling. To support a $360^\circ$ LiDAR, we divide the horizontal plane into four $90^\circ$ sectors (front, left, back, and right), each with an independent pyramid subspace division with a fixed orientation relative to the sensor frame. The DSP update is performed in parallel across the four sectors, providing full $360^\circ$ coverage.

\paragraph{Kalman Filter for Cluster Velocity Estimation}


The DSP map assigns initial velocities to newly generated particles based on the estimated velocity of the obstacle cluster to which they belong \cite{chen2023continuous}. The original method employs Euclidean cluster extraction using a $k$-d tree \cite{rusu2010semantic} and the Kuhn-Munkres (KM) algorithm for cluster matching. For matched clusters, obstacle velocity is estimated from cluster centroid displacement. However, direct position differences can provide poor velocity estimates because the cluster centroid is not a body-fixed point and can vary with viewing angle, obstacle rotation, and partial occlusion. Consequently, changes in cluster shape can be incorrectly interpreted as obstacle motion.

We modify the velocity estimation module to decouple variations in cluster shape from actual obstacle motion. Each cluster is characterized by its centroid $\mathbf{c}\in\mathbb{R}^2$ and scatter matrix $\bm{S}\in\mathbb{R}^{2\times2}$:
\begin{equation}
\mathbf{c}=\frac{1}{M_c}\sum_{i=1}^{M_c}\bm{p}_i,\qquad
\bm{S}=\sum_{i=1}^{M_c}(\bm{p}_i-\bm{c})(\bm{p}_i-\bm{c})^\top,
\label{eq:: Scatter}
\end{equation}
where $\bm{p}_i$ are the cluster points and $M_c$ is the number of points in the cluster.
For each pair of matched clusters, we employ a constant-velocity Kalman filter \cite{bar2001estimation} whose state consists of $x$ and $y$ obstacle position $(x^o,y^o)$ and velocity $(\dot{x}^o,\dot{y}^o)$ 
with the cluster centroid $\mathbf{c}$ serving as the position measurement. To reduce the influence of centroid shifts caused by shape changes, the measurement noise covariance $\bm{R}$ is adapted online. Specifically, we use a separate constant-acceleration filter to track the cluster's spatial spread and its temporal derivatives, providing a spread-aware measurement noise model.
The standard deviations of the cluster-point spread along the $x$ and $y$ directions are denoted by $\sigma_x$ and $\sigma_y$, respectively.
We define the spread state as \(\boldsymbol{\sigma} = [\sigma_x, \dot{\sigma}_x, \ddot{\sigma}_x, \sigma_y, \dot{\sigma}_y, \ddot{\sigma}_y]^\top \in \mathbb{R}^6\) and the measurements of $\sigma_x$ and $\sigma_y$ are given by $\mathbf{z}^\sigma = [\sqrt{\bm{S}_{00}},\; \sqrt{\bm{S}_{11}}]^\top$, where $\mathbf{S}$ is the scatter matrix defined in \eqref{eq:: Scatter}.
The estimated spread velocities, $[\hat{\dot{\sigma}}_x,\hat{\dot{\sigma}}_y]^\top$, obtained from the Kalman filter are used to modulate the measurement noise covariance of the constant-velocity Kalman filter for obstacle velocity estimation:
\begin{equation}\label{eq:kf_R_adaptive}
\bm{R} = \begin{bmatrix}
R_0 + \alpha|\hat{\dot{\sigma}}_{x}| & 0\\
0 & R_0 + \alpha|\hat{\dot{\sigma}}_{y}|
\end{bmatrix}
\end{equation}
with base noise \(R_0 = 0.001\) and modulation gain \(\alpha = 0.2\).
A rapidly expanding cluster (high \(\hat{\dot{\sigma}}_x, \hat{\dot{\sigma}}_y\)) indicates deformation or partial occlusion, increasing the measurement uncertainty and preventing the filter from following transient centroid jumps.
The final estimated velocity \(\hat{\bm{v}}_t = [\hat{\dot{x}}^o, \hat{\dot{y}}^o]^\top\) is used in place of the raw centroid displacement for particle propagation in the DSP-map prediction step, improving the quality of future occupancy predictions for dynamic obstacles.

\vspace{-0.3cm}
\subsection{Future Particle Occupancy Extraction}\label{sec:cloud_extraction}
The humanoid footstep plan is composed of alternate left and right single leg stance phases and the double stance phase is assumed to be absent. Each single stance phase has a duration of $T=0.3s$. We define the $k$th step of the humanoid over the interval $t\in[kT,(k+1)T)$.
We use the current particle states from the DSP map and a constant-velocity model to predict and accumulate the particles over the entire $k$th step. This accumulated particle cloud is denoted by $\hat{\mathcal{O}}_k$ and represents the space occupied by obstacles during the $k^{th}$ step. Given $t_0 \in [(k-1)T, kT]$ or in the $(k-1)^{th}$ step, we compute the accumulated particle clouds over the next $N+1$ prediction windows of length $T$, $\{\hat{\mathcal{O}}_{k}\}_{n=k}^{k+N}$, following the approach in \cite{chen2022rast}. Here N denotes the horizon of the MPC which performs footstep planning for the humanoid.
For particle accumulation within each window $n\in{k,k+1,..k+N}$, we propagate the particles to discrete evaluation points $\tau \in \{(n-k)T, (n-k)T + 0.1, (n-k)T + 0.2\}$ using a constant-velocity model,
\(
\tilde{\mathbf{p}}_{t_k + \tau}^{(i)} = \tilde{\mathbf{p}}_{t_0}^{(i)} + \tilde{\mathbf{v}}_{t_0}^{(i)}(\tau+t_{rem})
\)
Here, $t_{rem}$ is the time remaining in the current $(k-1)^{th}$ step at $t=t_0$, and $t_k = t_0 + t_{rem}$ is the predicted start time of the $k^{th}$ step. These evaluated particles are accumulated to give us the representative particle cloud for each window $n$
. Furthermore, we simplify the representation to have one particle at the center of each voxel with a weight equal to the sum of weights of all particles in that voxel. We further filter out voxels with low weights to give us the future occupancy particle clouds, $\{\tilde{\mathcal{O}}_{k}\}_{n=k}^{k+N}$ given by:

\begin{equation}\label{eq:cloud_extract}
\tilde{\mathcal{O}}_{k} = \Bigl\{\mathbf{b} \in \mathbb{R}^3 \;\Big|\; w^b_k > \rho_{\text{occ}}\Bigr\},
\end{equation}
where $\mathbf{b}$ is the voxel centroid, $w^b_k$ is the accumulated weight of the voxel and $\rho_{\text{occ}} = 0.1$ is the occupancy threshold. The clouds $\tilde{\mathcal{O}}_{k}$ are filtered via Euclidean clustering and flattened to 2D to produce noise-free clouds $\mathcal{O}_{k}$.

%% file: sections/03_methodology.tex
\section{Planning and Control}
\label{sec:planning&control}
The overall pipeline consists of the following sequential components: A* global path planning, LIP-based footstep sampling for seed extraction, iterative spatio-temporal safe polytope generation, desired velocity computation, LIP-MPC for footstep planning, and WBC for torque computation.
Figure \ref{fig:pipeline} illustrates the overall architecture.
\vspace{-0.3cm}
\subsection{A* Global Path Planning}\label{sec::Astar}
We utilize A* search \cite{hart1968formal} to compute a collision-free path for local goal guidance in downstream planners. We construct a binary occupancy grid from the LiDAR point cloud measurements. This grid is further used for planning a collision-free, low-cost path. We additionally penalize immediate local deviations from the previously computed path to avoid sudden changes in the direction of motion and fluctuations in the local plan.

\subsubsection{Grid Construction and Inflation}

Given the sensor position \(\mathbf{x}^s \in \mathbb{R}^2\), we construct a \(20 \times 20\)~m grid \(\mathcal{G}\) with a resolution of \(r_{\text{grid}} = 0.2\)~m, centered at \(\mathbf{x}^s\).
The grid is populated using the current LiDAR point cloud measurements, with a value of 1 assigned to occupied cells and 0 to free cells, thus forming a binary occupancy grid.
Furthermore, the occupied voxels in the grid are inflated by a safety margin of \(R_{\text{inflate}} = 1.0\)~m.

\subsubsection{A* Search and Local Goals}

An A* search finds the path \(\mathcal{P}_{\text{astar}}\) starting from the current CoM position, \(\mathbf{x}^c = \begin{bmatrix} x^c & y^c\end{bmatrix}^\top\) to the global goal, \(\bm{g}_{\text{global}}\) that minimizes the accumulated cell-to-cell path cost. We use the Euclidean distance to the goal as an admissible heuristic to guide the search.

When a previous path \(\mathcal{P}_{\text{prev}}\) exists, the A* cost is augmented with a deviation cost that biases the new path to stay close to \(\mathcal{P}_{\text{prev}}\) in the immediate vicinity of the robot.
For each candidate grid cell $i$, let  \(\mathbf{r}_i\in\mathbb{R}^2\) denote its center. We compute the deviation cost, $c_{dev}$ as: 
\begin{align}
c_{dev}(\mathbf{r}_i) &= c_0 \, d_{\perp}(\mathbf{r}_i) \, w\bigl(s(\mathbf{r}_i)\bigr),\nonumber\\
w(s) &= \begin{cases}
w_{\text{near}}, & s \leq L_{\text{lookahead}}\\
w_{\text{far}}, & \text{otherwise}
\end{cases}
\label{eq:deviation_cost}
\end{align}
Here, \(d_{\perp}(\mathbf{r}_i)\) is the perpendicular distance from the grid center, $\mathbf{r}_i$ to \(\mathcal{P}_{\text{prev}}\). If $\mathbf{p}_{\mathbf{r}_i}$ is the foot of the perpendicular from $\mathbf{r}_i$ on $\mathcal{P}_{prev}$, then \(s(\mathbf{r}_i)\) is the arc length of $\mathbf{p}_{\mathbf{r}_i}$ on \(\mathcal{P}_{\text{prev}}\). $c_0$ is a scalar constant and $L_\mathrm{lookahead}$ is the distance on the A* path for which we want to minimize the deviation from $\mathcal{P}_{prev}$. To ensure this, we set a higher value for $w_{near}$, while a smaller value is chosen for \(w_{far}\) to enable fast replanning. This deviation cost ensures that the A* path maintains consistency for the near future and avoids erratic changes in the direction of motion.
The accepted path is then smoothed via moving-average to remove grid-scale zigzags while preserving endpoints to obtain the path, $\tilde{\mathcal{P}}$.
Given $t_0\in[(k-1)T, kT]$ or in the $(k-1)^{th}$ step, we utilize $\tilde{\mathcal{P}}$ to obtain the goal points \(\mathbf{g}_i~\forall~i \in \{k, k+1, \dots, k+N\}\). These are calculated as \(N+1\) equidistant points along \(\tilde{\mathcal{P}}\) over the lookahead distance \(L_{\text{lookahead}} = 2.0\)~m, starting from the point on the path closest to \(\mathbf{x}^c\).
The farthest point \(\bm{g}_{k+N}\) provides global directional guidance, while the intermediate points refine the local heading for the downstream waypoint computation.
At each goal point, $\mathbf{g}_i$ the desired yaw for the stance foot of the humanoid is computed as the slope of the local path tangent at that point using central finite difference. We denote these yaw goals by $\theta^s_i~\forall~i\in\{k, k+1, \dots, k+N\}$, where the subscript $k$ denotes the goal for the $k^{th}$ step stance foot yaw.
%
%
\vspace{-0.3cm}
\subsection{Capture-Region-Constrained LIP Sampling}\label{sec:lip_sampling}
This section presents the sampling-based planner for generating dynamically and kinematically feasible, stable, and collision-free locomotion plans. The best obtained trajectory is used to extract seed points for subsequent safe polytope generation.
We use LIP dynamics to propagate the state and sample kinematically feasible footstep sequences over a horizon of $N$ footsteps, generating candidate CoM trajectories with $N+1$ states each. A key feature of the planner is the use of $M$-step capture regions\footnote{Terminology: A state is \emph{$M$-step capturable} if there exists an evolution from it that brings the system to a stop within $M$ steps, respecting dynamic and actuation constraints. The \emph{$M$-step capture basin} is the subset of states that are $M$-step capturable. An \emph{$M$-step capture point} is a ground location where a contact leads to an $(M-1)$-step capturable state; the set of all such points defines the \emph{$M$-step capture region}, $\mathcal{C}_M$; see~\cite{koolen2012capturability, Kumbhar2025Finite}.} to constrain footstep sampling. This ensures that all the sampled trajectories are stable, allowing a small number of samples to be used to extract dynamically and kinematically feasible seed points.

\subsubsection{Capture Region-based Trajectory Sampling}
We sample a total of $N_{pool}=1000$ candidate CoM trajectories using the LIP model dynamics. Let $\mathbf{u}_k = [u^x_k, u^y_k]^\top$, $\mathbf{x}^c_k = [x^c_k, y^c_k]^\top$ and $\dot{\mathbf{x}}^c_k = [\dot{x}^c_k, \dot{y}^c_k]^\top$ denote the x and y stance foot position, CoM position and velocity at the start of the $k^{th}$ step. The stance foot is assumed to remain the same during the entire step.
Each sampled candidate consists of a footstep sequence \(\{\mathbf{u}_i\}_{i=k}^{k+N-1}\) and a CoM trajectory \(\{\mathbf{x}^c_i,\dot{\mathbf{x}}^c_i\}_{i=k}^{k+N}\). 

\paragraph{Dynamic and Kinematic Feasibility}
We use the LIP model~\cite{kajita2001} for CoM propagation, with the discrete dynamics given by $\mathbf{X}^c_{k+1}=\mathbf{A}\mathbf{X}^c_k+\mathbf{B}\mathbf{u}_k$, where $\mathbf{X}^c_k=[x^c_k,\dot{x}^c_k,y^c_k,\dot{y}^c_k]^\top$. The analytical expression for matrices $\mathbf{A}$ and $\mathbf{B}$ can be found in \cite{Poulakakis2022Sequential}.
The discrete evolution of the system must adhere to the robot's kinematic constraints, which can be captured by the foot reachability set, $\mathcal{U}_k(\theta^\mathrm{s}_k, \mathbf{u}_k)$, parametrized by the stance foot position ($\mathbf{u}_k$) and yaw ($\theta^\mathrm{s}_k$) during the $k^{th}$ step. The sampled footstep for the $(k+1)^{th}$ step ($\mathbf{u}_{k+1}$) must belong to $\mathcal{U}(\theta^s_k, \mathbf{u}_k)$ for kinematic feasibility of the extracted seed points. This region is asymmetric in the lateral direction and its expression can be found in \cite{Kumbhar2025MPC}.
To simplify the sampling, we have $\theta^s_i = \theta^s_{k-1}~\forall~i\in\{k,k+1,...,k+N-1\}$, i.e., plan along the current heading direction. This guarantees that a plan to reach these seed points exists and that the MPC can fall back to refining this constant-yaw plan in case of failure.

\paragraph{Stable Trajectories}
We utilize the capturability framework developed in \cite{koolen2012capturability} to constrain foot placements such that the CoM evolution remains within the $M$-step capture basin, thereby ensuring a stable walking trajectory. The capture point ($\boldsymbol{\xi} = [\xi^x, \xi^y]^\top$) for LIP dynamics as defined by the capturability framework is given by,
\(
    \boldsymbol{\xi} = \mathbf{x}^c + \mathbf{\dot{x}}^c/\omega
    \).
Here, $\omega = \sqrt{g/H}$ and $H$ is the constant height at which the LIP CoM moves. Due to the decoupled and simple LIP dynamics, $M$-step capture region ($\mathcal{C}_M$) can be explicitly computed in the form, 
\(
    \mathcal{C}_M = \mathcal{C}^x_M \times \mathcal{C}^y_M
\)
for $M \in \{0,1,...\} \cup \{ \infty \}$. Here, the superscript $x$ and $y$ denote the $M$-step capture regions derived for the $x$ and $y$ axes. 
The sets $\mathcal{C}^x_M$ and  $\mathcal{C}^y_M$ are given by the following expressions
\vspace{-1mm}
\begin{align}
    \mathcal{C}^x_M &= \{r \in \mathbb{R} ~|~ \underline{d}^x_M \leq r \leq \overline{d}^x_M\}\\
    \mathcal{C}^y_M &= \{r \in \mathbb{R} ~|~ \underline{d}^y_M \leq r \leq \overline{d}^y_M \} 
\end{align}
The expressions for these upper and lower bounds have been analytically computed in \cite{Kumbhar2025Finite}. Constraining the capture point offset, defined as $\boldsymbol{\xi}_k - \mathbf{u}_k$, to lie within $\mathcal{C}_M$ as:
\vspace{-2mm}
\begin{align}
    \underline{d}^x_M \leq \xi^x_k-u^x_k \leq \overline{d}^x_M, \quad
    \underline{d}^y_M \leq \xi^y_k - u^y_k \leq \overline{d}^y_M \label{eq::capture},
\end{align}
guarantees that the footsteps produce stable and $M$-step capturable CoM trajectories.
Using the conditions in \eqref{eq::capture} 
, we further constrain the footsteps to reduce the sampling space to
\vspace{-4mm}
\begin{align}
u^x_k &\in  [\xi^x_k + \underline{d}^x_{M^x_k},\; \xi_x^k + \overline{d}^x_{M^x_k}]\cap \mathcal{U}^x_k, \label{eq::footsampleX}\\
u^y_k &\in [\xi^y_k + \underline{d}^y_{M^x_k},\; \xi_y^k + \overline{d}^y_{M^x_k}] \cap \mathcal{U}^y_k, \label{eq::footsampleY}
\end{align}
Here, $\mathcal{U}^x_k$ and $\mathcal{U}^y_k$ denote the kinematic reachability sets for the forward and lateral directions of motion \cite{Poulakakis2022Sequential}: $M^x_k = max(M^x_0, M^x{-}k-1)$, $M^y_k = max(M^y_0, M^y{-}k-1)$, where
$M^x$ and $M^y$ denote the minimum number of steps required for the current state to be capturable in the $x-$ and $y-$directions, respectively, while
$M^x_0$ and $M^y_0$ specify the minimum $M$-step capture basin for the CoM state during sampling. The capture regions are computed such that the set resulting from the intersections in \eqref{eq::footsampleX} and \eqref{eq::footsampleY} is guaranteed to be non-empty. Consequently, ensuring that all $N_{pool}$ samples exist and are stable, dynamically, and kinematically feasible CoM trajectories. The initial state for planning is obtained by propagating the current robot state to the end of the stance phase using LIP dynamics. 
We utilize the discrete step dynamics of LIP and the above sampled footstep locations to generate candidate CoM trajectories ($\mathcal{T}$).

\subsubsection{Collision Checking}

The $k^{th}$ segment of LIP is defined as the CoM trajectory evolving from $\mathbf{x}^c_k$ to $\mathbf{x}^c_{k+1}$ with $\mathbf{u}_k$ as the stance foot. A sampled trajectory ($\mathcal{T}$) consists of N steps and so N segments of CoM evolution. Using the analytical solutions of the LIP dynamics, we can compute a bounding box $\mathcal{B}_k$ that encompasses all the CoM positions in the $k^{th}$ segment trajectory. For collision checking, we compute such bounding boxes for all segments of the sampled trajectory and also inflate the boundaries by a safety margin defined by $\delta_b$.

A segment is collision free if it satisfies the following condition:
\begin{equation}
    (x,y) \notin \mathcal{B}_k \quad \forall (x,y) \in \mathcal{O}_k
    \label{eq:: collisioncheck}
\end{equation}
where $\mathcal{O}_k$ is the flattened future occupancy map as computed in Section \ref{sec:cloud_extraction}. We discard all the candidate CoM trajectories that do not satisfy the collision check given by \eqref{eq:: collisioncheck}.

\subsubsection{Trajectory Cost}

Each non-colliding trajectory is evaluated by the cost
\vspace{-0.3cm}
\begin{equation}\label{eq:cost}
J = 1.5\bigl\|\mathbf{x}^c_{k+N-1} - \mathbf{g}_{k+N}\bigr\| +
0.5\sum_{i=k}^{k+N-2} \bigl\|\mathbf{x}^c_{i} - \mathbf{g}_{k+N}\bigr\|,
\end{equation}
weighting the terminal CoM position more heavily to steer the robot toward the local goal, $\mathbf{g}_{k+N}$.
The trajectory with the lowest cost among all collision free candidates is retained.

\subsubsection{Seed Position Extraction}

The \(N+1\) CoM positions (\(\{\mathbf{x}^c_i\}_{i=k}^{k+N}\)) are extracted from the best trajectory as seed points (\(\{\bm{s}_i\}_{i=0}^{N}\)) for safe polytope generation.

\vspace{-0.3cm}
\subsection{Iterative Spatio-Temporal Safe-Polytope Generation}

Given the future occupancy particle cloud $(\{\mathcal{O}_i\}^{k+N}_{i=k})$ from Section \ref{sec:cloud_extraction} and seed points from the sampling planner, we compute convex collision-free regions, $\{\mathcal{S}_i\}^{k+N}_{i=k}$. $\mathcal{S}_k$ is a convex polytope that is considered safe for navigation during the $k^{th}$ step. We propose an iterative risk-based algorithm for computing structurally smooth and large polytopes. Algorithm \P{\ref{algo1}} is provided below. First, we explain the general polytope construction method for a set of objects, then describe our iterative algorithm.

\subsubsection{Polytope Construction via Separating Half-planes}

The function \(\textsc{ComputePolytope}(\mathcal{O}, \mathbf{s})\) constructs a convex safe polytope from a set of 2D obstacle positions \(\mathcal{O}\) and a seed point \(\mathbf{s}\).
A safe polytope \(\mathcal{S}\) is the intersection of half-plane constraints:
\begin{equation}\label{eq:polytope_def}
\mathcal{S} = \{\bm{p} \in \mathbb{R}^2 \mid \bm{a}_j^\top \bm{p} \leq b_j,\; j = 1, \dots, m\}.
\end{equation}
Here, $\mathbf{a}^T_j$ is the slope of the $j^{th}$ line and $b_j$ is the corresponding offset.
To construct this polytope, obstacles are first sorted by ascending distance from the seed point, \(\bm{s}\). For each obstacle, if its Cartesian position, \(\mathbf{o}\in \mathcal{O}\), already lies behind an existing half-plane (\(\bm{a}_j^\top \bm{o} > b_j\) for some \(j\)), it is skipped. 
Otherwise, a separating half-plane given by $\bm{n}^\top \bm{p} \leq \bm{n}^\top \mathbf{o}$ is added with its boundary on the obstacle, where $\bm{n}$ is the outward normal given by $\bm{n} = (\mathbf{o} - \mathbf{s}) / |\mathbf{o} - \mathbf{s}|$.
To ensure that the planned CoM trajectory starting from the current seed point to the next also lies inside the polytope $\mathcal{S}$, we require that the bounding box containing this trajectory, $\mathcal{B}$, lies in $\mathcal{S}$. Therefore, after the construction of a half-plane, we identify the corner points of $\mathcal{B}$ that lie on the other side of the plane as the seed point. Then, for each corner point, we form the line joining the obstacle point $\mathbf{o}$ and the corner point. We select the line that keeps the entire box and the seed point on the same side. In the event of a tie break, we select the normal that is closest to the original normal $\bm{n}$. This ensures that the bounding box is on the safe side the the hyperplane.
The returned polytope is the intersection of all such constraints. It is guaranteed to contain the bounding box and consequently, the entire planned trajectory between seed points.
The obstacle for which we computed the plane becomes an \emph{anchor}, one that actively constrains the polytope. $\textsc{FINDANCHORS}(\mathcal{S})$ function stores a list of these anchor points that produce the closed polytope\footnote{The polytope computation starts with a box constraint around the seed point and as a result a closed polytope is always possible.}, $\mathcal{S}$.

\subsubsection{Iterative Weighted Expansion with Per-Cluster Thresholding}

The DSP-map particle clouds are inherently noisy, with particle additions and deletions causing temporal variations in particle representation of the obstacle shape. Treating each particle as an obstacle during polytope construction can therefore produce abrupt changes in polytope size and shape, which may reduce the feasibility of downstream footstep planning. We propose an algorithm that selectively drops anchor particles (those generating active constraints) given an acceptable risk of collision with the particles. Each particle $\mathbf{o}_j\in\mathcal{O}_k$ has weight $w_j$ and cluster identifier $c_j$. Let $\mathcal{O}^w_k=\{(\mathbf{o}_j,w_j,c_j)\}$ and $W_c=\sum_{j:c_j=c}w_j$ denote the total weight of cluster $c$. We define the acceptable risk of collision ($\eta$) as the ratio of the total weight of particles inside the safe polytope to the total weight of all particles in the map. Given an allowable collision risk $\eta\in [0~1]$, we assign each cluster a weight-removal threshold
$T_c=\eta W_c$. This threshold acts as a limit on how many anchor particles can be dropped from a cluster while ensuring the risk does not exceed $\eta$.
Thus, the cumulative weight removed from any cluster cannot exceed its allocated risk budget.

The algorithm starts with computing the initial polytope $\mathcal{S}$ and the anchor points $\mathcal{A}$.
For each of these anchors $\mathbf{a}\in\mathcal{A}$, we compute a polytope for the set of all particles except the anchor itself. 
We denote these polytopes as candidate polytopes and represent them by $\mathcal{S}_a$. The corresponding anchors for these polytopes are denoted by $\mathcal{A}_a$. For each candidate, we compute a lexicographic cost function defined below:
\begin{equation}\label{eq:anchor_cost}
\text{cost}(\mathbf{a}) = \big(|\mathcal{A}_{\bm{a}}|,\;
-\sum_{\mathbf{a}' \in \mathcal{A}_{\mathbf{a}}^{}} d(\mathbf{a}'),\;
w(\mathbf{a})\big)
\end{equation}
Here, $w(\mathbf{a}$) is the weight of the anchor $\mathbf{a}$ and $d(\mathbf{a}')$ is the shortest distance between the seed and the plane corresponding to anchor $\mathbf{a}'$.
This cost prioritizes: (1) minimizing remaining anchors, (2) maximizing the sum of anchor distances from the seed point, (3) dropping lower-weight obstacles first.
The first priority objective of minimizing the number of anchors prevents sudden shape changes because of outlier particles which can be ignored under acceptable risk. The second priority objective ensures that the polytope boundaries are further away from the seed points giving us larger area for safe planning. The third priority objective ensures risk minimization. 

Candidates are sorted lexicographically by \eqref{eq:anchor_cost} in ascending order.
The first candidate whose cluster's dropped weight would not exceed its threshold after dropping is selected:
\begin{equation}\label{eq:cluster_check}
D_c + w(\mathbf{a}) \leq T_c,
\end{equation}
where \(D_c\) is the cumulative weight already dropped from cluster \(c\).
The dropped weight is then updated: \(D_c \gets D_c + w(\bm{a})\) and the polytopes and anchors for the next iteration are computed by dropping the selected anchor.
If no candidate satisfies \eqref{eq:cluster_check}, the loop terminates.
This is repeated independently for each of the \(N+1\) prediction windows to get $\{\mathcal{S}_i\}^{k+N}_{i=k}$.
The per-cluster threshold prevents the algorithm from removing too many safety constraints from any single obstacle cluster.
This is critical in cluttered scenes where a single dynamic obstacle generates many points near the robot. Without a per-cluster limit, those points could all be dropped, leaving the region unsafe.
\vspace{-0.45cm}
\subsection{Velocity Command Extraction \& LIP-MPC}
Given the A*-derived lookahead points \(\{\mathbf{g}_i\}_{i=k}^{k+N}\) and the safe polytopes \(\{\mathcal{S}_i\}_{i=k}^{k+N}\), we compute smooth CoM waypoints \(\{\mathbf{w}_i\}_{i=1}^{N+1}\) for desired velocity computation via a Quadratic Program as follows:
\begin{align}
\min_{\bm{w}_1,\dots,\bm{w}_{N+1}}\;
&\sum_{i=1}^{N+1} \gamma^s_i \|\mathbf{w}_i - \mathbf{w}_{i-1}\|^2
+ \sum_{i=1}^{N+1} \gamma_g \|\mathbf{w}_i - \mathbf{g}_{k+i-1}\|^2
\label{eq:qp_cost}
\end{align}
\vspace{-0.5cm}
\begin{align}
\mathrm{subject\hspace{0.1cm} to}\hspace{0.29cm}&\mathbf{w}_i \in \mathcal{S}_{k+i-1}\cap\mathcal{S}_{k+i},\quad i = 1,\dots,N \label{eq:qp_constr1}\\
&\mathbf{w}_{N+1} \in \mathcal{S}_{k+N} \label{eq:qp_constr2}
\end{align}
where \(\mathbf{w}_0 = \mathbf{x}^{c}\) is the current CoM position and $\gamma^s_i~\forall~i\in\{1,2,..,N+1\}$, $\gamma_g$ are positive scalars.
The first term penalizes deviation from the current position, the second encourages consecutive waypoints to be close, and the third attracts each waypoint toward its A*-derived lookahead goal.
Constraint \eqref{eq:qp_constr1} enforces that each waypoint lies in the intersection of consecutive polytopes, enabling safe transitions of the robot between polytopes.
Desired velocities for the LIP-MPC are computed from waypoint differences as
\(
\mathbf{v}_{\text{des}}^{k+i} = (\mathbf{w}_{i+1} - \mathbf{w}_i)/T,\quad i = 1,\dots,N.
\)
These velocity commands and the desired yaw computed in Section \ref{sec::Astar} are passed as targets to the MPC for footstep planning.
The resulting convex MPC formulation is given as follows:
\vspace{0.06in}
\hrule 

\vspace{-0.06in}
\begin{align}
    &\operatorname*{minimize}_{\mathbf{X}^\mathrm{c}_{i+1}, \mathbf{u}_i}\quad \sum_{i=k}^{k+N-1} J_i\big(\mathbf{X}^\mathrm{c}_{i},\mathbf{u}_{i-1}, \theta^\mathrm{s}_{i-1}\big)\\
    &\mathrm{subject\hspace{0.1cm} to}\hspace{0.29cm}\mathbf{X}^c_{i+1} = \mathbf{A} \mathbf{X}^c_i + \mathbf{B} \mathbf{u}_i\label{ILIP_eq}\\
    &\hspace{1.9cm}\mathbf{u}_i=(u^x_i,u^y_i)\in \mathcal{U}_i(\theta^\mathrm{s}_{i-1}, \mathbf{u}_{i-1})\label{foot_bound_eq}\\
    &\hspace{1.9cm}\mathbf{x}^c_i\in \mathcal{S}_i ~\forall i\in\{k,k+1,..k+N-1\}\label{pol_1}\\
    &\hspace{1.9cm}\mathbf{x}^c_{i+1}\in \mathcal{S}_i ~\forall i\in\{k,k+1,..k+N-2\}\label{pol_2}
\end{align}
\vspace{-0.18in}
\hrule
\vspace{0.2cm}
The decision variables of the optimization are the robot’s CoM states, \(\mathbf{X}^\mathrm{c}_{i+1}\), and the stance foot positions, \(\mathbf{u}_{i}\), for all \(i \in \{k, k+1, \ldots, k+N-1\}\).
The optimization is subject to following constraints:  
(i) the step-to-step dynamics of the LIP model~\eqref{ILIP_eq},
(ii) the foot placement constraint~\eqref{foot_bound_eq}, and (iii) CoM state belongs to the intersection of consecutive safe polytopes~\eqref{pol_1},~\eqref{pol_2}. 
The MPC objective ($J_i)$ combines two terms: a goal-velocity tracking cost and a CoM-foot distance regulation cost. The former penalizes the deviation between the CoM velocity and desired velocity $\mathbf{v}^i_{des}$, while the latter penalizes the CoM displacement from the stance foot, discouraging dynamically unfavorable footstep placements. The planned footsteps and CoM trajectory are then provided to the whole-body controller (WBC), which generates the corresponding torque commands. Details of the WBC can be found in \cite{Poulakakis2022Sequential}.

\begin{algorithm}[t]
\small
\caption{Iterative Spatio-temporal Safe-Polytope Generation}
\label{alg:polytope}
\begin{algorithmic}[1]
\State $W_c \gets \sum_{j: c_j = c} w_j$,\; $T_c \gets \eta W_c$,\; $D_c \gets 0$ for each cluster $c$
\State $\mathcal{S} \gets \textsc{ComputePolytope}(\mathcal{O}_k,\mathbf{s}_k)$
\State $\mathcal{A} \gets \textsc{FindAnchors}(\mathcal{S})$
\While{$\mathcal{A} \neq \varnothing$}
\State $\mathcal{C} \gets \varnothing$
\For{each $\bm{a} \in \mathcal{A}$}
\State $\mathcal{S}_{\bm{a}} \gets \textsc{ComputePolytope}(\mathcal{O}\setminus\{\bm{a}\},s)$
\State $\mathcal{A}_{\bm{a}} \gets \textsc{FindAnchors}(\mathcal{S}_{\bm{a}})$
\State $\mathcal{C} \gets \mathcal{C} \cup \{(\bm{a},\; \text{cost}(\bm{a}))\}$
\EndFor
\State sort $\mathcal{C}$ by cost in ascending order
\State $\bm{a}^* \gets \varnothing$
\For{$(\bm{a},\cdot) \in \mathcal{C}$}
\If{$D_{c(\bm{a})} + w(\bm{a}) \leq T_{c(\bm{a})}$}
\State $\bm{a}^* \gets \bm{a}$; \Break
\EndIf
\EndFor
\If{$\bm{a}^* = \varnothing$}
\State \Break
\EndIf
\State $D_{c(\bm{a}^*)} \gets D_{c(\bm{a}^*)} + w(\bm{a}^*)$
\State $\mathcal{S} \gets \textsc{ComputePolytope}(\mathcal{O}\setminus\{\bm{a}^*\},s)$
\State $\mathcal{A} \gets \textsc{FindAnchors}(\mathcal{S})$
\State $\mathcal{O} \gets \{\mathcal{O}\setminus\{\bm{a}^*\}\}$
\EndWhile
\State \Return $\mathcal{S}$
\end{algorithmic}
\label{algo1}
\end{algorithm}

%% file: sections/04_experiments.tex
\section{Results}
\label{sec:results}
We evaluate the proposed framework in dynamic environments containing static and dynamic non-convex obstacles and human actors performing random walks \footnote{A video of the simulation demonstrations is available at: \\https://www.youtube.com/watch?v=nuLPCccjv84}. The humanoid is tasked with navigating to a goal located within a moving U-shaped object. This goal-containing object and other dynamic obstacles move at constant velocities. Unlike approaches that approximate non-convex obstacles using convex representations, the DSP-map-based framework does not impose a predefined obstacle geometry, enabling navigation both around and within non-convex obstacles. In all experiments, the goal position within the non-convex region is assumed to be available in the humanoid's local frame, decoupling high-level semantic goal specification from motion planning and control.

\begin{figure*}[t]
    \centering
    \includegraphics[width=0.98\textwidth]{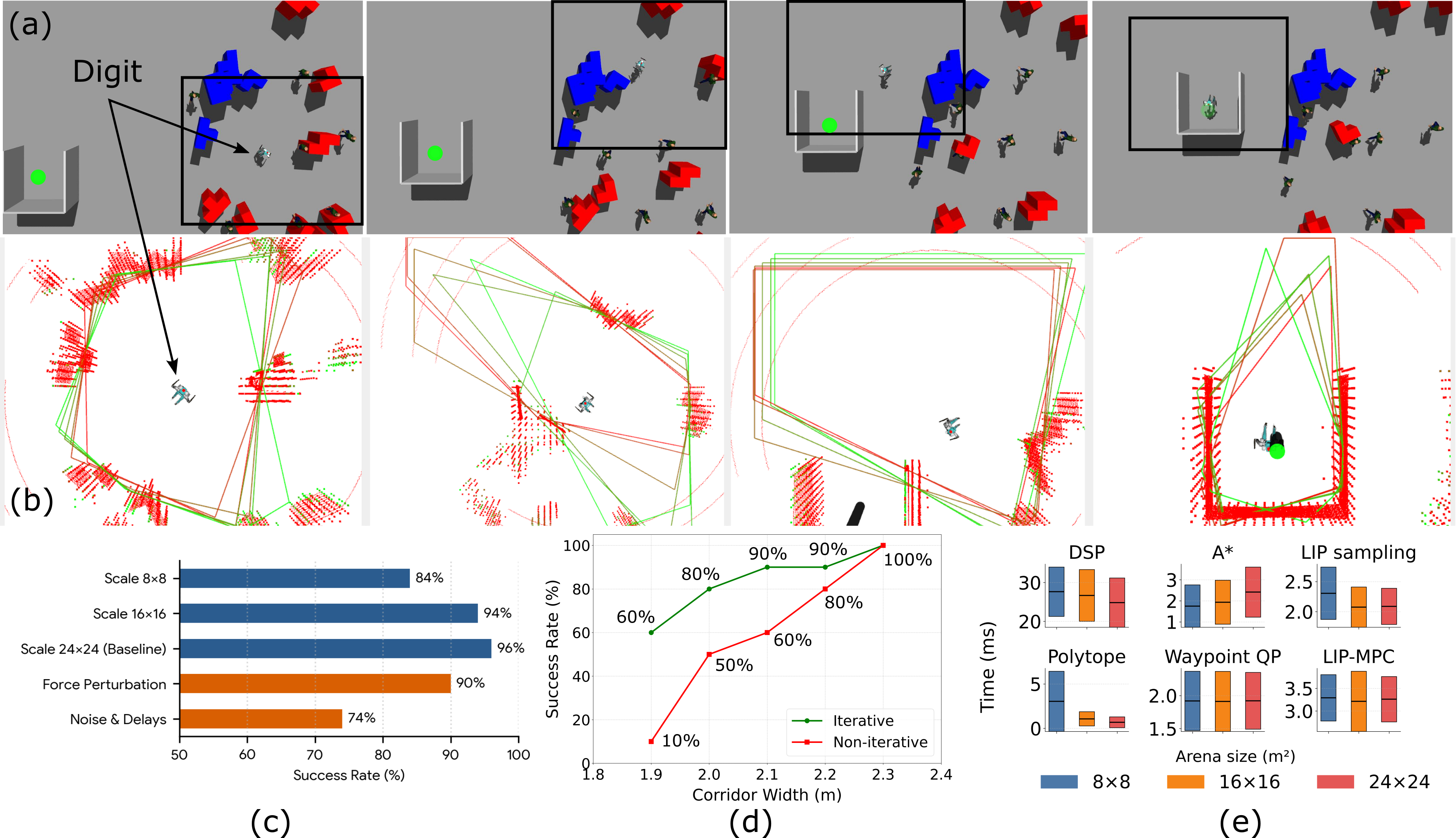}
    \vspace{-0.35cm}
    \caption{(a) Snapshots of the Digit robot navigating in an $8\times8~\mathrm{m}^2$ arena containing human actors, moving non-convex obstacles (red), static obstacles (blue), and a U-shaped obstacle (gray), with the goal located at its center. (b) Corresponding safe corridors generated by the proposed algorithm and the future occupancy maps for the regions highlighted by the black boxes in (a). Five prediction windows are illustrated, with the first shown in green and the color gradually transitioning to red for the fifth prediction window. The green circle represents the goal position inside the gray-colored U-shaped moving object. (c) Success rates for three arena scales, force perturbations, measurement noise, and delays. (d) Success rates for varying narrow-corridor widths for both iterative and non-iterative cases. (e) Computation time of each module for a successful experiment in each of the three arena scales. The center of each bar represents the mean computation time, while the upper and lower extents indicate one standard deviation.
    \vspace{-0.5cm}
    }
    \label{fig:results}
\end{figure*}

Experiments are conducted in MuJoCo \cite{todorov2012mujoco} using Agility Robotics' Digit humanoid~\cite{digit}, equipped with a Velodyne VLP-16 LiDAR. LiDAR observations are generated at a frequency of 20Hz through kinematic simulation of Digit in Gazebo \cite{koenig2004design}, with Gaussian measurement noise, control delays, and motor dynamics incorporated to better reflect real-world operation. Figure~\ref{fig:results}(a) shows representative snapshots of the humanoid navigating successfully among static and dynamic non-convex obstacles, along with the generated spatio-temporal safe corridors shown in Fig.~\ref{fig:results}(b).
\vspace{-0.3cm}
\subsection{Large-Scale Experimentation}
We evaluate the framework in randomly generated environments containing L-, T-, and Z-shaped non-convex obstacles and human actors. The numbers of static and dynamic obstacles and human actors are independently sampled from 3--7 each. Human actors perform random walks with a speed of $0.8$~m/s, while dynamic obstacles and the U-shaped goal object are assigned random directions of motion and speeds between $-0.5$ and $0.5$~m/s. A trial is successful if the humanoid reaches the goal without collision or falling. A fall is declared when the torso height drops below $0.5$~m or the roll/pitch magnitude exceeds $0.785$~rad, while collisions are determined from the distance between the robot CoM and obstacle wall segments. We perform the following two evaluations:

\subsubsection{Arena Scale and Clutter Density Evaluation}

We evaluate three arena sizes, $24\times24~\mathrm{m}^2$, $16\times16~\mathrm{m}^2$, and $8\times8~\mathrm{m}^2$, with the same obstacle-count range. The arena area decreases by a factor of nine from the largest to the smallest case, while the obstacle footprint remains unchanged, resulting in substantially higher clutter density in the $8\times8~\mathrm{m}^2$ arena.
We conduct 50 trials for each case. As shown in Fig.~\ref{fig:results}(c), the framework maintains a high success rate in the two larger arenas, with performance degrading in the $8\times8$~m arena as the fixed obstacle footprint leaves less maneuvering space and requires more aggressive footstep placement. 
Most failures are falls rather than collisions, indicating that obstacle avoidance is maintained but the humanoid increasingly approaches its kinematic and dynamic limits. Additionally, the lack of predictive capabilities in the global guidance provided by $A^*$ can lead the robot to narrow spaces that could otherwise be avoided.
\subsubsection{Evaluation under Sim-to-Real Imperfections}

We evaluate robustness to external perturbations by applying random forces on the robot's torso every $2$~s, uniformly sampled from $[-50,50]$~N and applied for $100$~ms \cite{Poulakakis2022Sequential}. As shown in Fig.~\ref{fig:results}(c), the success rate for 50 trials in the $24\times24~m^2$ arena remains comparable to the baseline case, with failures primarily caused by pushes that leave no capturable safe foot placement.
Robustness to noise and a conservative $15$~ms control delay is evaluated using empirical parameters from~\cite{kumbhar2025efficient}, with $\sigma=0.01$ for joint position/velocity ($\mathrm{rad}$, $\mathrm{rad/s}$) and torso position/velocity/orientation/angular velocity ($\mathrm{m}$, $\mathrm{m/s}$, $\mathrm{rad}$, $\mathrm{rad/s}$). The framework achieves a $74\%$ success rate over 50 trials in the $24\times24~m^2$ (see Fig.~\ref{fig:results}(c)), demonstrating robustness to moderate sensing and actuation imperfections. 
\vspace{-0.4cm}
\subsection{Narrow Corridor: Iterative vs. Non-Iterative Polytope Generation}


We conduct an ablation study in $5$~m-long corridors of varying widths to evaluate risk-aware iterative polytope generation. We compare the proposed iterative method with a single polytope-generation iteration as a non-iterative baseline. Ten trials are conducted per corridor width, and success is defined as reaching the opposite end without collision or falling. As shown in Fig.~\ref{fig:results}(d), both methods perform comparably in wide corridors, but the non-iterative method degrades sharply as the corridor narrows. With the conservative $1$~m robot-size assumption, the iterative method successfully traverses corridors as narrow as $1.9$~m.
The non-iterative baseline retains low-weight particles as active constraints, resulting in more half-space constraints and unnecessarily smaller safe regions. Moreover, its polytopes exhibit abrupt temporal shape changes when the robot is near obstacles. In contrast, the iterative method selectively removes anchors while respecting the cluster-level risk threshold and minimizes the number of active half-spaces per obstacle. This produces larger and more consistent free regions, enabling successful navigation through narrower regions.

        


\vspace{-0.3cm}
\subsection{Computational Performance}
Figure~\ref{fig:results}(e) summarizes the per-module computation times for successful experiments across the three arena scales. The DSP-map generation and planning and control modules run in separate threads, enabling fast motion replanning. For all experiments in the results section, 1000 LIP trajectories are sampled for seed-point extraction. The low sampling time is due to the reduced sampling space imposed by the capture-region constraints.
The computation times for the planning and control modules remain small and relatively consistent across the experimental conditions, with a modest increase in polytope computation in the $8\times8$~m arena due to its higher obstacle density. We also observe a modest increase in DSP-map update time as clutter density increases. The complete system runs on a desktop equipped with an Intel i9 CPU and 64~GB of RAM, without GPU acceleration.




%% file: sections/05_conclusion.tex
\section{Conclusion}
\label{sec:conclusion}
This paper presented an integrated, real-time perception-to-control framework for humanoid navigation in dynamic environments with non-convex obstacles. The architecture bridges $360^\circ$ LiDAR perception and locomotion control by unifying a modified DSP map, $A^*$ global guidance, capture-region-constrained LIP sampling, iterative safe corridor generation, LIP-MPC, and WBC.
The key algorithmic contribution is a spatio-temporal polytope construction that combines capture-region-constrained LIP sampling with an iterative polytope generation algorithm. By pruning dynamically unrecoverable states early via $M$-step capture regions, the sampler efficiently extracts feasible seed points with few samples. The iterative polytope generation algorithm uses these seed points to construct spatially and temporally consistent convex free regions for future times without restrictive shape assumptions.
Extensive simulation experiments on the Digit humanoid demonstrate a 96\% success rate for the $24~m\times24~m$ arena, validating the framework's ability to navigate both around and inside moving non-convex obstacles.
Ablations confirm the iterative polytope's superiority in narrow corridors, while evaluations show resilience to sensor noise, delays, and force perturbations. Future work will target predictive global planning and hardware validation.